\documentclass{article}

\usepackage{agi-2010}
\usepackage{graphicx}
\usepackage{amsmath}
\usepackage{amssymb}
\usepackage{amsthm}
\usepackage{psfrag}
\usepackage{booktabs}
\usepackage{natbib}

\newcommand{\fs}[1]{\mathcal{#1}}

\newcommand{\define}{\equiv}

\newcommand{\comment}[1]{}

\newcommand{\finint}{\fs{Z}^\ast}   
\newcommand{\infint}{\fs{Z}^\infty} 
\newcommand{\prob}{\mathbf{P\!r}}      

\newcommand{\g}[1]{\underline{#1}}

\newcommand{\agent}{\mathbf{P}}
\newcommand{\env}{\mathbf{Q}}
\newcommand{\bagent}{\tilde{\mathbf{P}}}
\newcommand{\gen}{\mathbf{G}}

\theoremstyle{plain}

\theoremstyle{definition}

\begin{document}

\bibliographystyle{plainnat}

\title{A Bayesian Rule for Adaptive Control based on Causal Interventions}

\author{Pedro A. Ortega\\
        Department of Engineering\\
        University of Cambridge\\
        Cambridge CB2 1PZ, UK\\
        \texttt{peortega@dcc.uchile.cl}\\
        \And
        Daniel A. Braun\\
        Department of Engineering\\
        University of Cambridge\\
        Cambridge CB2 1PZ, UK\\
        \texttt{dab54@cam.ac.uk}\\
}


\maketitle

\begin{abstract}%
Explaining adaptive behavior is a central problem in artificial intelligence
research. Here we formalize adaptive agents as mixture distributions over
sequences of inputs and outputs (I/O). Each distribution of the mixture
constitutes a `possible world', but the agent does not know which of the
possible worlds it is actually facing. The problem is to adapt the I/O stream
in a way that is compatible with the true world. A natural measure of
adaptation can be obtained by the Kullback-Leibler (KL) divergence between the I/O
distribution of the true world and the I/O distribution expected by the agent
that is uncertain about possible worlds. In the case of pure input streams, the
Bayesian mixture provides a well-known solution for this problem. We show,
however, that in the case of I/O streams this solution breaks down, because
outputs are issued by the agent itself and require a different probabilistic
syntax as provided by intervention calculus. Based on this calculus, we obtain
a Bayesian control rule that allows modeling adaptive behavior with mixture
distributions over I/O streams. This rule might allow
for a novel approach to adaptive control based on a minimum KL-principle.
\end{abstract}


\section{Introduction}
The ability to adapt to unknown environments is often considered a hallmark of
intelligence \citep{Beer1990,Hutter2004}. Agent and environment can be
conceptualized as two systems that exchange symbols in every time step
\citep{Hutter2004}: the symbol issued by the agent is an action, whereas the
symbol issued by the environment is an observation. Thus, both agent and
environment can be conceptualized as probability distributions over sequences
of actions and observations (I/O streams).

If the environment is perfectly known then the I/O probability distribution of
the agent can be tailored to suit this particular environment. However, if the
environment is unknown, but known to belong to a set of possible environments,
then the agent faces an adaptation problem. Consider, for example, a robot that
has been endowed with a set of behavioral primitives and now faces the problem
of how to act while being ignorant as to which is the correct primitive. Since
we want to model both agent and environment as probability distributions over
I/O sequences, a natural way to measure the degree of adaptation would be to
measure the `distance' in probability space between the I/O distribution
represented by the agent and the I/O distribution conditioned on the true
environment. A suitable measure (in terms of its information-theoretic
interpretation) is readily provided by the KL-divergence \citep{Mackay2003}. In
the case of passive prediction, the adaptation problem has a well-known
solution. The distribution that minimizes the KL-divergence is a Bayesian
mixture distribution over all possible environments \citep{Opper1997,
Opper1998}. The aim of this paper is to extend this result for distributions
over both inputs and outputs. The main result of this paper is that this
extension is only possible if we consider the special syntax of actions in
probability theory as it has been suggested by proponents of causal calculus
\citep{Pearl2000}.

\section{Preliminaries}

We restrict the exposition to the case of discrete time with discrete
stochastic observations and control signals. Let $\fs{O}$ and $\fs{A}$ be two
finite sets, the first being the \emph{set of observations} and the second
being the \emph{set of actions}. We use $a_{\leq t} \define a_1 a_2 \ldots
a_t$, $\g{ao}_{\leq t}
\define a_1 o_1 \ldots a_t o_t$ etc. to simplify the notation of
strings. Using $\fs{A}$ and $\fs{O}$, a set of interaction sequences is
constructed. Define the \emph{set of interactions} as $\fs{Z} \define \fs{A}
\times \fs{O}$. A pair $(a,o) \in \fs{Z}$ is called an \emph{interaction}. The
set of interaction strings of length $t \geq 0$ is denoted by $\fs{Z}^t$.
Similarly, the set of (finite) interaction strings is $\finint
\define \bigcup_{t \geq 0} \fs{Z}^t$ and the set of (infinite) interaction sequences
is $\infint \define \{ w: w = a_1 o_1 a_2 o_2 \ldots \}$, where each $(a_t,
o_t) \in \fs{Z}$. The interaction string of length 0 is denoted by $\epsilon$.

Agents and environments are formalized as I/O systems. An \emph{I/O system} is
a probability distribution $\prob$ over interaction sequences $\infint$.
$\prob$ is uniquely determined by the conditional probabilities
\begin{equation}\label{eq:cs-stream}
    \prob(a_t|\g{ao}_{<t}), \quad \prob(o_t|\g{ao}_{<t}a_t)
\end{equation}
for each $\g{ao}_{\leq t} \in \finint$. However,
the semantics of the probability distribution $\prob$ are only fully defined
once it is coupled to another system.

Let $\agent$, $\env$ be two I/O systems. An \emph{interaction system}
$(\agent, \env)$ is a coupling of the two systems giving rise to
the \emph{generative distribution} $\gen$ that describes the probabilities
that actually govern the I/O stream once the two systems are coupled. $\gen$ is
specified by the equations
\begin{align*}
    \gen(a_t|\g{ao}_{<t}) &= \agent(a_t|\g{ao}_{<t}) \\
    \gen(o_t|\g{ao}_{<t}a_t) &= \env(o_t|\g{ao}_{<t}a_t)
\end{align*}
valid for all $\g{ao}_t \in \finint$. Here, $\gen$ models the true probability
distribution over interaction sequences that arises by coupling two
systems through their I/O streams. More specifically, for the system $\agent$,
$\agent(a_t|\g{ao}_{<t})$ is the probability of producing action $a_t \in
\fs{A}$ given history $\g{ao}_{<t}$ and $\agent(o_t|\g{ao}_{<t}a_t)$ is the
predicted probability of the observation $o_t \in \fs{O}$ given history
$\g{ao}_{<t}a_t$. Hence, for $\agent$, the sequence $o_1 o_2 \ldots$ is its
input stream and the sequence $a_1 a_2 \ldots$ is its output stream. In
contrast, the roles of actions and observations are reversed in the case of the
system $\env$. Thus, the sequence $o_1 o_2 \ldots$ is its output
stream and the sequence $a_1 a_2 \ldots$ is its input stream. This model of
interaction is very general in that it can accommodate many specific regimes of
interaction. Note that an agent $\agent$ can perfectly predict its
environment $\env$ iff for all $\g{ao}_{\leq t} \in \finint$,
\[
    \agent(o_t|\g{ao}_{<t}a_t) = \env(o_t|\g{ao}_{<t}a_t).
\]
In this case we say that $\agent$ is \emph{tailored} to $\env$.

\section{Adaptive Systems: Na\"{\i}ve Construction}

Throughout this paper, we use the convention that $\agent$ is an \emph{agent}
to be constructed by a designer, which is then going to be interfaced with a
preexisting but unknown \emph{environment} $\env$. The designer assumes that
$\env$ is going to be drawn with probability $P(m)$ from a set $\fs{Q}
\define \{ \env_m \}_{m \in \fs{M}}$ of possible systems before the interaction
starts, where $\fs{M}$ is a countable set.

Consider the case when the designer knows beforehand which environment $\env
\in \fs{Q}$ is going to be drawn. Then, not only can $\agent$ be tailored to
$\env$, but also a custom-made policy for $\env$ can be designed. That is, the
output stream $\agent(a_t|\g{ao}_{<t})$ is such that the true probability
$\gen$ of the resulting interaction system $(\agent,\env)$ gives rise to
interaction sequences that the designer considers \emph{desirable}.

Consider now the case when the designer does not know which environment $\env_m
\in \fs{Q}$ is going to be drawn, and assume he has a set $\fs{P} \define \{
\agent_m \}_{m \in \fs{M}}$ of  systems such that for each $m \in \fs{M}$,
$\agent_m$ is tailored to $\env_m$ and the interaction system $(\agent_m,
\env_m)$ has a generative distribution $\gen_m$ that produces desirable
interaction sequences. How can the designer construct a
 system $\agent$ such that its behavior is as close as possible to
the custom-made  system $\agent_m$ under any realization of $\env_m
\in \fs{Q}$?

A convenient measure of how much $\agent$ deviates from $\agent_m$ is given by
the KL-divergence. A first approach would be to construct an agent $\bagent$
so as to minimize the total expected KL-divergence to $\agent_m$. This
is constructed as follows. Define the history-dependent KL-divergences over
the action $a_t$ and observation $o_t$ as
\begin{align*}
    D_m^{a_t}(\g{ao}_{<t}) &\define
        \sum_{a_t} \agent_m(a_t|\g{ao}_{<t})
        \log_2 \frac{ \agent_m(a_t|\g{ao}_{<t}) }{ \prob(a_t|\g{ao}_{<t}) }
        \\
    D_m^{o_t}(\g{ao}_{<t}a_t) &\define
        \sum_{o_t} \agent_m(o_t|\g{ao}_{<t}a_t)
        \log_2 \frac{ \agent_m(o_t|\g{ao}_{<t}a_t) }{ \prob(o_t|\g{ao}_{<t}a_t)
        },
\end{align*}
where $\prob$ is a given arbitrary  agent. Then, define the average
KL-divergences over $a_t$ and $o_t$ as
\begin{align*}
    D_m^{a_t} &= \sum_{\g{ao}_{<t}}
        \agent_m(\g{ao}_{<t}) D_m^{a_t}(\g{ao}_{<t}) \\
    D_m^{o_t} &= \sum_{\g{ao}_{<t}a_t}
        \agent_m(\g{ao}_{<t}a_t) D_m^{o_t}(\g{ao}_{<t}a_t).
\end{align*}
Finally, we define the total expected KL-divergence of $\prob$ to $\agent_m$
as
\[
    D \define \limsup_{t \rightarrow \infty}
        \sum_{m} P(m) \sum_{\tau=1}^t \bigl(
            D_m^{a_\tau} + D_m^{o_\tau} \bigr).
\]
We construct the agent $\bagent$ as the  system that minimizes $D =
D(\prob)$:
\begin{equation}\label{eq:minimization-bayesian}
    \bagent \define \arg \min_{\prob} D(\prob).
\end{equation}
The solution to Equation~\ref{eq:minimization-bayesian} is the  system
$\bagent$ defined by the set of equations
\begin{equation}\label{eq:bayesian-agent}
\begin{aligned}
    \bagent(a_t|\g{ao}_{<t})
        &= \sum_m \agent_m(a_t|\g{ao}_{<t}) w_m(\g{ao}_{<t})\\
    \bagent(o_t|\g{ao}_{<t}a_t)
        &= \sum_m \agent_m(o_t|\g{ao}_{<t}a_t) w_m(\g{ao}_{<t}a_t)
\end{aligned}
\end{equation}
valid for all $\g{ao}_{\leq t} \in \finint$, where the mixture weights are
\begin{equation}\label{eq:bayesian-weights}
\begin{aligned}
    w_m(\g{ao}_{<t}) &\define
        \frac{ P(m) \agent_m(\g{ao}_{<t}) }
             { \sum_{m'} P(m') \agent_{m'}(\g{ao}_{<t}) }
    \\
    w_m(\g{ao}_{<t}a_t) &\define
        \frac{ P(m) \agent_m(\g{ao}_{<t}a_t) }
             { \sum_{m'} P(m') \agent_{m'}(\g{ao}_{<t}a_t) }.
\end{aligned}
\end{equation}
For reference, see \citet{Opper1997, Opper1998}. It is clear that $\bagent$ is
just the Bayesian mixture over the
 agents $\agent_m$. If we define the conditional probabilities
\begin{equation}\label{eq:op-mode-streams}
\begin{aligned}
    P(a_t|m,\g{ao}_{<t}) &\define \agent_m(a_t|\g{ao}_{<t}) \\
    P(o_t|m,\g{ao}_{<t}a_t) &\define \agent_m(a_t|\g{ao}_{<t}a_t)
\end{aligned}
\end{equation}
for all $\g{ao}_{\leq t} \in \finint$, then Equation~\ref{eq:bayesian-agent}
can be rewritten as
\begin{equation}\label{eq:bayesian-agent-2}
\begin{aligned}
    \bagent(a_t|\g{ao}_{<t})
        &= \sum_m P(a_t|m,\g{ao}_{<t}) P(m|\g{ao}_{<t})\\
    \bagent(o_t|\g{ao}_{<t}a_t)
        &= \sum_m P(o_t|m,\g{ao}_{<t}a_t) P(m|\g{ao}_{<t}a_t)
\end{aligned}
\end{equation}
where the $P(m|\g{ao}_{<t})=w_m(\g{ao}_{<t})$ and
$P(m|\g{ao}_{<t}a_t)=w_m(\g{ao}_{<t}a_t)$ are just the posterior probabilities
over the elements in $\fs{M}$ given the past interactions. Hence, the
conditional probabilities in Equation~\ref{eq:op-mode-streams}, together with
the prior probabilities $P(m)$, define a Bayesian model over interaction
sequences with hypotheses $m \in \fs{M}$.

The behavior of $\bagent$ can be described as follows. At any given time $t$,
$\bagent$ maintains a mixture over  systems $\agent_m$. The weighting over them
is given by the mixture coefficients $w_m$. Whenever a new action $a_t$
\emph{or} a new observation is produced (by the agent or the environment
respectively), the weights $w_m$ are updated according to Bayes' rule. In
addition, $\bagent$ issues an action $a_t$ suggested by a system $\agent_m$
drawn randomly according to the weights $w_t$.

However, there is an important problem with $\bagent$ that arises due to the fact
that it is not only a system that is passively observing symbols, but also
\emph{actively generating} them. Therefore, an action that is generated by
the agent should not provide the same information than an observation that
is issued by its environment. Intuitively, it does not make any sense
to use one's own actions to do inference. In the following section
we illustrate this problem with a simple statistical example.

\section{The Problem of Causal Intervention}

Suppose a statistician is asked to design a model for a given data set $\fs{D}$
and she decides to use a Bayesian method. She computes the posterior
probability density function (pdf) over the parameters $\theta$ of the model
given the data using Bayes' rule:
\[
    p(\theta|\fs{D}) =
    \frac{ p(\fs{D}|\theta) p(\theta) }
         { \int p(\fs{D}|\theta') p(\theta') \, d\theta' },
\]
where $p(\fs{D}|\theta)$ is the likelihood of $\fs{D}$ given $\theta$ and
$p(\theta)$ is the prior pdf of $\theta$. She can simulate the source by
drawing a sample data set $\fs{S}$ from the predictive pdf
\[
    p(\fs{S}|\fs{D})
    = \int p(\fs{S}|\fs{D},\theta) p(\theta|\fs{D}) \, d\theta,
\]
where $p(\fs{S}|\fs{D},\theta)$ is the likelihood of $\fs{S}$ given $\fs{D}$
and $\theta$. She decides to do so, obtaining a sample set $\fs{S}'$. She
understands that the nature of $\fs{S}'$ is very different from $\fs{D}$:
\emph{while $\fs{D}$ is informative and does change the belief state of the
Bayesian model, $\fs{S}'$ is non-informative and thus is a reflection of the
model's belief state.} Hence, she would never use $\fs{S}'$ to further
condition the Bayesian model. Mathematically, she seems to imply that
\[
    p(\theta|\fs{D},\fs{S}') = p(\theta|\fs{D})
\]
if $\fs{S}'$ has been generated from $p(\fs{S}|\fs{D})$ itself. But this simple
independence assumption is not correct as the following elaboration of the
example will show.

The statistician is now told that the source is waiting for the simulation
results $\fs{S}'$ in order to produce a next data set $\fs{D}'$ which does
depend on $\fs{S}'$. She hands in $\fs{S}'$ and obtains a new data set
$\fs{D}'$. Using Bayes' rule, the posterior pdf over the parameters is now
\begin{equation}\label{eq:intervened-posterior}
    \frac{ p(\fs{D}'|\fs{D},\fs{S}',\theta)
           p(\fs{D}|\theta) p(\theta) }
         { \int p(\fs{D}'|\fs{D},\fs{S}',\theta')
           p(\fs{D}|\theta') p(\theta') \, d\theta' }
\end{equation}
where $p(\fs{D}'|\fs{D},\fs{S}',\theta)$ is the likelihood of the new data
$\fs{D}'$ given the old data $\fs{D}$, the parameters $\theta$ \emph{and the
simulated data} $\fs{S}'$. Notice that this looks almost like the posterior pdf
$p(\theta|\fs{D},\fs{S}',\fs{D}')$ given by
\[
    \frac{ p(\fs{D}'|\fs{D},\fs{S}',\theta)
           p(\fs{S}'|\fs{D},\theta)
           p(\fs{D}|\theta) p(\theta) }
         { \int p(\fs{D}'|\fs{D},\fs{S}',\theta')
           p(\fs{S}'|\fs{D},\theta')
           p(\fs{D}|\theta') p(\theta') \, d\theta' }
\]
with the exception that now the Bayesian update contains the likelihoods of the
simulated data $p(\fs{S}'|\fs{D},\theta)$. This suggests that
Equation~\ref{eq:intervened-posterior} is a variant of the posterior pdf
$p(\theta|\fs{D},\fs{S}',\fs{D}')$ but where the simulated data $\fs{S}'$ is
treated in a different way than the data~$\fs{D}$ and~$\fs{D}'$.

Define the pdf $p'$ such that the pdfs $p'(\theta)$, $p'(\fs{D}|\theta)$,
$p'(\fs{D}'|\fs{D},\fs{S}',\theta)$ are identical to $p(\theta)$,
$p(\fs{D}|\theta)$ and $p(\fs{D}'|\fs{D},\fs{S}',\theta)$ respectively, but
differ in $p'(\fs{S}|\fs{D},\theta)$:
\[
    p'(\fs{S}|\fs{D},\theta) =
    \begin{cases}
        1 & \text{if $\fs{S}' = \fs{S}$,} \\
        0 & \text{else.}
    \end{cases}
\]
That is, $p'$ is identical to $p$ but it assumes that the value of $\fs{S}$ is
fixed to $\fs{S}'$ given $\fs{D}$ and $\theta$. For $p'$, the simulated data
$\fs{S}'$ is non-informative:
\[
    -\log_2 p(\fs{S}'|\fs{D},\theta) = 0.
\]
If one computes the posterior pdf $p'(\theta|\fs{D},\fs{S}',\fs{D}')$, one
obtains the result of Equation~\ref{eq:intervened-posterior}:
\begin{align*}
    \frac{ p'(\fs{D}'|\fs{D},\fs{S}',\theta)
           p'(\fs{S}'|\fs{D},\theta)
           p'(\fs{D}|\theta) p'(\theta) }
         { \int p'(\fs{D}'|\fs{D},\fs{S}',\theta')
           p'(\fs{S}'|\fs{D},\theta')
           p'(\fs{D}|\theta') p'(\theta') \, d\theta' }
    \\
    = \frac{ p(\fs{D}'|\fs{D},\fs{S}',\theta)
             p(\fs{D}|\theta) p(\theta) }
           { \int p(\fs{D}'|\fs{D},\fs{S}',\theta')
             p(\fs{D}|\theta') p(\theta') \, d\theta' }.
\end{align*}
Thus, in order to explain Equation~\ref{eq:intervened-posterior} as a posterior
pdf given the data sets $\fs{D}$, $\fs{D}'$ and the simulated data $\fs{S}'$,
one has to \emph{intervene} $p$ in order to account for the fact that
\emph{$\fs{S}'$ is non-informative given $\fs{D}$ and $\theta$.}

\begin{figure*}[htbp]
\begin{center}
    \includegraphics[]{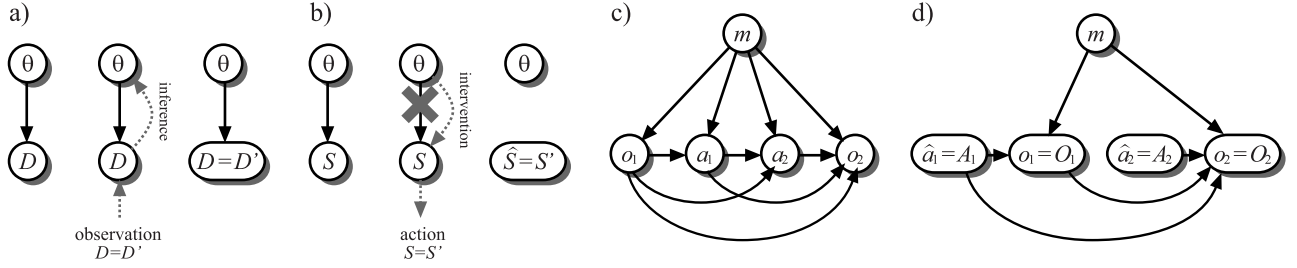}
    \caption{(a-b) Two causal networks, and the result of conditioning on
    $D = D'$ and intervening on $S = S'$. Unlike the condition,
    the intervention is set endogenously, thus removing the link
    to the parent $\theta$.
    (c-d) A causal network representation of an I/O system with
    four variables $a_1 o_1 a_2 o_2$ and latent variable $m$.
    (c) The initial, un-intervened network. (d) The intervened network
    after experiencing $\hat{a}_1 o_1 \hat{a}_2 o_2$.}
    \label{fig:agent}
\end{center}
\end{figure*}

In statistics, there is a rich literature on causal intervention. In
particular, we will use the formalism developed by \citet{Pearl2000}, because
it suits the needs to formalize interactions in  systems and has a convenient
notation---compare Figures \ref{fig:agent}a \&~b. Given a \emph{causal
model}\footnote{For our needs, it is enough to think about a causal model as a
complete factorization of a probability distribution into conditional
probability distributions representing the causal structure.} variables that
are intervened are denoted by a hat as in $\hat{\fs{S}}$. In the previous
example, the causal model of the joint pdf $p(\theta, \fs{D}, \fs{S}, \fs{D}')$
is given by the set of conditional pdfs
\[
    \fs{C}_p = \bigl\{
       p(\theta), p(\fs{D}|\theta),
       p(\fs{S}|\fs{D},\theta),
       p(\fs{D}'|\fs{D},\fs{S},\theta)
    \bigr\}.
\]
If $\fs{D}$ and $\fs{D}'$ are observed from the source and $\fs{S}$ is
intervened to take on the value $\fs{S}'$, then the posterior pdf over the
parameters $\theta$ is given by $p(\theta|\fs{D},\hat{\fs{S}'},\fs{D}')$ which
is just
\begin{align*}
    \frac{ p(\fs{D}'|\fs{D},\hat{\fs{S}'},\theta)
           p(\hat{\fs{S}'}|\fs{D},\theta)
           p(\fs{D}|\theta) p(\theta) }
         { \int p(\fs{D}'|\fs{D},\hat{\fs{S}'},\theta')
           p(\hat{\fs{S}'}|\fs{D},\theta')
           p(\fs{D}|\theta') p(\theta') \, d\theta' }
    \\
    = \frac{ p(\fs{D}'|\fs{D},\fs{S}',\theta)
             p(\fs{D}|\theta) p(\theta) }
           { \int p(\fs{D}'|\fs{D},\fs{S}',\theta')
             p(\fs{D}|\theta') p(\theta') \, d\theta' }.
\end{align*}
because $p(\fs{D}'|\fs{D},\hat{\fs{S}'},\theta) =
p(\fs{D}'|\fs{D},\fs{S}',\theta)$, which corresponds to applying rule~2
in Pearl's intervention calculus,
and because $p(\hat{\fs{S}'}|\fs{D},\theta') = p'(\fs{S}'|\fs{D},\theta') = 1$.

\section{Adaptive Systems: Causal Construction}

Following the discussion in the previous section, we want to construct an
adaptive agent $\agent$ by minimizing the KL-divergence to the $\agent_m$, but
this time treating actions as interventions. Based on the definition of the
conditional probabilities in Equation~\ref{eq:op-mode-streams}, we construct
now the KL-divergence criterion to characterize $\agent$ using intervention
calculus. Importantly, interventions index a set of intervened probability
distribution derived from an initial probability distribution. Hence, the set
of fixed intervention sequences of the form $\hat{a}_1 \hat{a}_2 \ldots$
indexes probability distributions over observation sequences $o_1 o_2 \ldots$.
Because of this, we are going to construct a set of criteria indexed by the
intervention sequences, but we will see that they all have the same solution.
Define the history-dependent intervened KL-divergences over the action $a_t$
and observation $o_t$ as
\begin{align*}
    C_m^{a_t}(\g{\hat{a}o}_{<t}) &\define
        \sum_{a_t} P(a_t|m,\g{\hat{a}o}_{<t})
        \log_2 \frac{ P(a_t|m,\g{\hat{a}o}_{<t}) }{ \prob(a_t|\g{ao}_{<t}) }
        \\
    C_m^{o_t}(\g{\hat{a}o}_{<t}\hat{a}_t) &\define
        \sum_{o_t} P(o_t|m,\g{\hat{a}o}_{<t}\hat{a}_t)
        \log_2 \frac{ P(o_t|m,\g{\hat{a}o}_{<t}\hat{a}_t) }{ \prob(o_t|\g{ao}_{<t}a_t)
        },
\end{align*}
where $\prob$ is a given arbitrary  agent. Note that past actions are
treated as interventions. Then, define the average KL-divergences over
$a_t$ and $o_t$ as
\begin{align*}
    C_m^{a_t} &= \sum_{\g{ao}_{<t}}
        P(\g{\hat{a}o}_{<t}|m) C_m^{a_t}(\g{\hat{a}o}_{<t}) \\
    C_m^{o_t} &= \sum_{\g{ao}_{<t}a_t}
        P(\g{\hat{a}o}_{<t}a_t|m) C_m^{o_t}(\g{\hat{a}o}_{<t}\hat{a}_t).
\end{align*}
Finally, we define the total expected KL-divergence of $\agent$ to
$\agent_m$ as
\begin{equation}
\label{eq:totalKL}
    C \define \limsup_{t \rightarrow \infty}
        \sum_{m} P(m) \sum_{\tau=1}^t \bigl(
            C_m^{a_\tau} + C_m^{o_\tau} \bigr).
\end{equation}
We construct the agent $\agent$ as the  system that minimizes $C =
C(\prob)$:
\begin{equation}\label{eq:minimization-causal}
    \agent \define \arg \min_{\prob} C(\prob).
\end{equation}
The solution to Equation~\ref{eq:minimization-causal} is the  system
$\agent$ defined by the set of equations
\begin{equation}\label{eq:causal-agent}
\begin{aligned}
    \agent(a_t|\g{ao}_{<t})
        &= P(a_t|\g{\hat{a}o}_{<t}) \\
        &= \sum_m P(a_t|m,\g{\hat{a}o}_{<t}) v_m(\g{\hat{a}o}_{<t})\\
    \agent(o_t|\g{ao}_{<t}a_t)
        &= P(o_t|\g{\hat{a}o}_{<t}\hat{a}_t) \\
        &= \sum_m P(o_t|m,\g{\hat{a}o}_{<t}\hat{a}_t) v_m(\g{\hat{a}o}_{<t}\hat{a}_t)
\end{aligned}
\end{equation}
valid for all $\g{ao}_{\leq t} \in \finint$, where the mixture weights are
\begin{equation}\label{eq:causal-weights}
\begin{aligned}
    v_m(\g{ao}_{<t}a_t)
    &= v_m(\g{ao}_{<t})
    \define \frac{ P(m) P(\g{\hat{a}o}_{<t}|m) }
                 { \sum_{m'} P(m') P(\g{\hat{a}o}_{<t}|m) }
    \\&= \frac{ P(m) \prod_{\tau=1}^{t-1}
                P(o_\tau|m,\g{\hat{a}o}_{<\tau}\hat{a}_\tau) }
            { \sum_{m'} P(m') \prod_{\tau=1}^{t-1}
                P(o_\tau|m',\g{\hat{a}o}_{<\tau}\hat{a}_\tau) }.
\end{aligned}
\end{equation}

The proof follows the same line of argument as the solution to
Equation~\ref{eq:minimization-bayesian} with the crucial difference that
actions are treated as interventions. Consider without loss of generality the
summand $\sum_m P(m) C^{a_t}_m$ in Equation~\ref{eq:totalKL}. Note that the
KL-divergence can be written as a difference of two logarithms, where only one
term depends on $\prob$ that we want to vary. Therefore, we can integrate out
the other term and write it as a constant $c$. Then we get
\begin{align*}
    c - \sum_{m} \, P(m)
    &\sum_{\g{\hat ao}_{<t}} \, P(\g{\hat ao}_{<t}|m) \,\,
    \\  &\cdot
        \sum_{a_t} P(a_t|m,\g{\hat ao}_{<t})
        \ln \prob(a_t|\g{\hat ao}_{<t}).
\end{align*}
Substituting $P(\g{\hat ao}_{<t}|m)$ by $P(m|\g{\hat ao}_{<t}) P(\g{\hat
ao}_{<t}) / P(m)$ and identifying $\agent$ characterized by
Equations~\ref{eq:causal-agent} and~\ref{eq:causal-weights} we obtain
\[
    c - \sum_{\g{\hat ao}_{<t}} \, P(\g{\hat ao}_{<t}) \,\,
        \sum_{a_t} \agent(a_t|\g{\hat ao}_{<t})
        \ln \prob(a_t|\g{\hat ao}_{<t}).
\]
The inner sum has the form $-\sum_x p(x) \ln q(x)$, i.e. the cross-entropy
between $q(x)$ and $p(x)$, which is minimized when $q(x)=p(x)$ for all $x$. By
choosing this optimum one obtains $\prob(a_t|\g{\hat ao}_{<t}) =
\agent(a_t|\g{\hat ao}_{<t})$ for all $a_t$. Note that the solution to this
variational problem is independent of the weighting $P(\g{\hat{a}o}_{<t})$.
Since the same argument applies to any summand $\sum_m P(m) C^{a_\tau}_m$ and
$\sum_m P(m) C^{o_\tau}_m$ in Equation~\ref{eq:totalKL}, their variational
problems are mutually independent.

The behavior of $\agent$ differs in an important aspect from $\bagent$. At any
given time $t$, $\agent$ maintains a mixture over  systems $\agent_m$. The
weighting over these systems is given by the mixture coefficients $v_m$. In
contrast to $\bagent$, $\agent$ updates the weights $v_m$ \emph{only} whenever
a new observation $o_t$ is produced by the environment respectively. The update
follows Bayes' rule but treating past actions as interventions, i.e. dropping
the evidence they provide. In addition, $\agent$ issues an action $a_t$
suggested by an system $m$ drawn randomly according to the weights~$v_m$---see
Figures~\ref{fig:agent}c \&~d.

If we use the following equalities connecting the weights and the intervened posterior
distributions
\[
    v_m(\g{ao}_{<t})
    = P(m|\g{\hat{a}o}_{<t})
    = P(m|\g{\hat{a}o}_{<t}\hat{a}_t)
    = v_m(\g{ao}_{<t}a_t)
\]
and substitute interventions by observations in the conditionals
\begin{align*}
    P(a_t|m,\g{\hat{a}o}_{<t})
    &= P(a_t|m,\g{ao}_{<t})
    \\P(o_t|m,\g{\hat{a}o}_{<t}\hat{a}_t)
    &= P(o_t|m,\g{ao}_{<t}a_t)
\end{align*}
which corresponds to rule~2 of Pearl's intervention calculus, we can rewrite
Equations~\ref{eq:causal-agent} and~\ref{eq:causal-weights} as
\begin{align}
    \nonumber
    \agent(a_t|\g{ao}_{<t})
        &= P(a_t|\g{\hat{a}o}_{<t}) \\
        \label{eq:bcr}
        &= \sum_m P(a_t|m,\g{ao}_{<t}) P(m|\g{\hat{a}o}_{<t})\\
    \nonumber
    \agent(o_t|\g{ao}_{<t}a_t)
        &= P(o_t|\g{\hat{a}o}_{<t}\hat{a}_t) \\
        \label{eq:bcr-obs}
        &= \sum_m P(o_t|m,\g{ao}_{<t}a_t) P(m|\g{\hat{a}o}_{<t})
\end{align}
where the intervened posterior probabilities are
\begin{equation}\label{eq:bcr-weights}
\begin{aligned}
    P(m|\g{\hat{a}o}_{<t})
    &= \frac{ P(m) \prod_{\tau=1}^{t-1}
                P(o_\tau|m,\g{ao}_{<\tau}a_\tau) }
            { \sum_{m'} P(m') \prod_{\tau=1}^{t-1}
                P(o_\tau|m',\g{ao}_{<\tau}a_\tau) }.
\end{aligned}
\end{equation}
Equations~\ref{eq:bcr}, \ref{eq:bcr-obs} and~\ref{eq:bcr-weights} are important
because they describe the behavior of $\agent$ only in terms of known
probabilities, i.e. probabilities that are computable from the causal model
associated to $P$ given by
\[
    C_P = \bigl\{
        P(m), P(a_t|m,\g{ao}_{<t}), P(o_t|m,\g{ao}_{<t}a_t): t \geq 1
        \bigr\}.
\]
Importantly, Equation~\ref{eq:bcr} describes a stochastic method to produce
desirable actions that differs fundamentally from an agent that is constructed
by choosing an optimal policy with respect to a given utility criterion. We
call this action selection rule the \emph{Bayesian control rule}.

\section{Experimental Results}

\begin{figure*}
\label{fig:simulation}
\centering
\includegraphics[width=7cm]{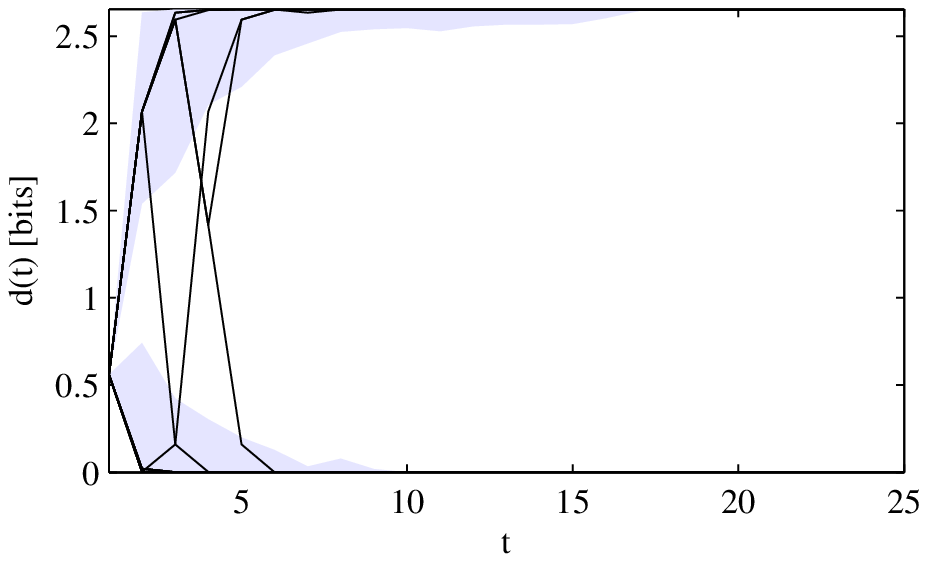}
\includegraphics[width=7cm]{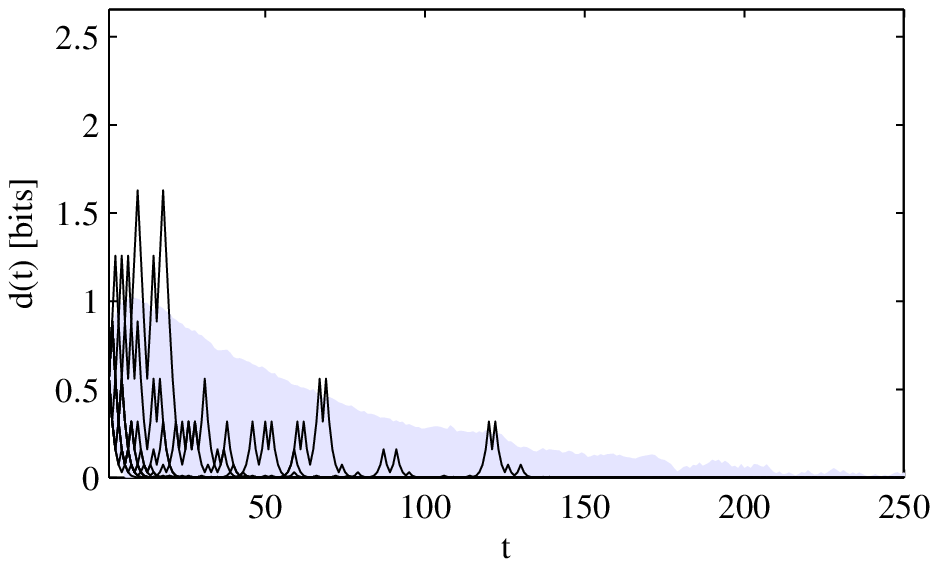}
\caption{10 realizations of the instantaneous deviation $d(t)$ for the agents
$\bagent$ (left panel) and $\agent$ (right panel). The shaded region represents
the standard deviation barriers computed over 1000 realizations. Since $d(t)$
is non-ergodic for $\bagent$, we have separated the realizations converging to
$0$ from the realizations converging to $\approx 2.654$ to compute the
barriers. Note that the time scales differ in one order of magnitude.}
\end{figure*}

Here we design a very simple toy experiment to illustrate the behavior of an agent $\bagent$
based on a Bayesian mixture compared to an agent $\agent$ based on the Bayesian
control rule.

Let $\env_0$, $\env_1$, $\agent_0$ and $\agent_1$ be four  agents
with binary I/O sets $\fs{A}=\fs{O}=\{0,1\}$ defined as follows. $\agent_1$ is
such that $\agent_1(a_t|\g{ao}_{<t}) = \agent_1(a_t)$ and
$\agent_1(o_t|\g{ao}_{<t}a_t) = \agent_1(o_t)$ for all $\g{ao}_{\leq t} \in
\finint$, where
\begin{equation*}
    \agent_1(a_t) =
        \begin{cases}
            0.1 & \text{if $a_t = 0$} \\
            0.9 & \text{if $a_t = 1$}
        \end{cases},\quad
    \agent_1(o_t) =
        \begin{cases}
            0.4 & \text{if $a_t = 0$} \\
            0.6 & \text{if $a_t = 1$}
        \end{cases}.
\end{equation*}
Let $\agent_0$ be such that
\begin{align*}
    \agent_0(a_t|\g{ao}_{<t}) &= 1 - \agent_1(a_t|\g{ao}_{<t}) \\
    \agent_0(o_t|\g{ao}_{<t}a_t) &= 1 - \agent_0(o_t|\g{ao}_{<t}a_t)
\end{align*}
for all $\g{ao}_{\leq t} \in \finint$. Thus, $\agent_0$ and $\agent_1$ are
 agents that are biased towards observing and acting 0's and 1's
respectively. Furthermore, $\env_0 = \agent_0$ and $\env_1 = \agent_1$. Assume
a uniform distribution over $\fs{Q} = \{\env_0, \env_1\}$, i.e. $P(m=0) =
P(m=1) = \frac{1}{2}$.

Assume $\env_0 \in \fs{Q}$ is drawn. In this case, one wants the agents
$\bagent$ and $\agent$ to minimize the deviation from $\agent_0$. Consider the
following instantaneous measure
\begin{align*}
    d(t)
    &\define \sum_{a_t'} \agent_0(a_t') \log_2
        \frac{ \agent_0(a_t') }{ \prob(a_t'|\g{ao}_{<t}) }
    \\&+ \sum_{o_t'} \agent_0(o_t') \log_2
        \frac{ \agent_0(o_t') }{ \prob(o_t'|\g{ao}_{<t}a_t) }
\end{align*}
where $a_1 o_1 a_2 o_2 \ldots$ is a realization of the interaction system
$(\prob, \env_0)$. $d(t)$ measures how much $\prob$'s action and observation
probabilities deviate from $\agent_0$ at time~$t$.

Recall that both $\bagent$ and $\agent$ maintain a mixture over $\agent_0$ and
$\agent_1$. The instantaneous I/O probabilities of such a  system can
always be written as
\begin{align*}
    &w \agent_0(a_t) + (1-w) \agent_1(a_t) \\
    &w \agent_0(o_t) + (1-w) \agent_1(o_t).
\end{align*}
where $w \in [0,1]$. Thus, it is easy to see that the instantaneous I/O
deviation takes on the minimum value when $w=1$ and the maximum value when
$w=0$: In the case $w=1$, $d(t) = 0$ bits; In the case $w=0$, $d(t) \approx 2.
653$.

We have simulated realizations of the instantaneous I/O deviation using the
agents $\bagent$ and $\agent$. The results are summarized in
Figure~2. For $\bagent$, $d(t)$ happens to be non-ergodic:
it either converges to $d(t) \rightarrow 0$ or to $d(t) \rightarrow \approx
2.654$, implying that either $\bagent \rightarrow \agent_0$ or $\bagent
\rightarrow \agent_1$ respectively. In contrast, $d(t) \rightarrow 0$ always
for $\agent$, implying that $\agent \rightarrow \agent_0$.

Analogous results are obtained when $\env_1 \in \fs{Q}$ is drawn instead: For
$\bagent$, $d(t)$ converges either to $0$ or to $\approx 2.654$, whereas for
$\agent$, $d(t) \rightarrow \approx 2.654$ always implying that $\agent
\rightarrow \agent_1$. Hence, $\agent$ shows the correct adaptive behavior
while $\bagent$ does not.

\section{Conclusions}\label{sec:conclusions}

We propose a Bayesian rule for adaptive control. The key feature of this rule
is the special treatment of actions based on causal calculus and the
decomposition of agents into Bayesian mixture of I/O distributions. The
question of how to integrate information generated by an agent's probabilistic
model into the agent's information state lies at the very heart of adaptive
agent design. We show that the na\"{\i}ve application of Bayes' rule to I/O
distributions leads to inconsistencies, because outputs don't provide the same
type of information as genuine observations. Crucially, these inconsistencies
vanish if intervention calculus is applied \citep{Pearl2000}.

Some of the presented key ideas are not unique to the Bayesian control rule.
The idea of representing agents and environments as I/O streams has been
proposed by a number of other approaches, such as predictive state
representation (PSR) \citep{Littman2002} and the universal AI approach by
\citet{Hutter2004}. The idea of breaking down a control problem into a
superposition of controllers has been previously evoked in the context of
``mixture of experts''-models like the MOSAIC-architecture \cite{Haruno2001}.
Other stochastic action selection approaches are found in exploration
strategies for (PO)MDPs \citep{Wyatt1997}, learning automata
\citep{Narendra1974} and in probability matching \citep{DudaHartStork2000}
amongst others. The usage of compression principles to select actions has been
proposed by AI researchers, for example \citet{Schmidhuber2009}. The main
contribution of this paper is the derivation of a stochastic action selection
and inference rule by minimizing KL-divergences of intervened I/O
distributions.

An important potential application of the Bayesian control rule would naturally
be the realm of adaptive control problems. Since it takes on a similar form to
Bayes' rule, the adaptive control problem could then be translated into an
on-line inference problem where actions are sampled stochastically from a
posterior distribution. It is important to note, however, that the problem
statement as formulated here and the usual Bayes-optimal approach in adaptive
control are \emph{not} the same. In the future the relationship between these
two problem statements deserves further investigation.

\vskip 0.2in
\bibliographystyle{unsrt}
\small
\bibliography{bibliography}

\end{document}